\documentclass[conference]{IEEEtran}
\IEEEoverridecommandlockouts
\usepackage{cite}
\usepackage{amsmath,amssymb,amsfonts}
\usepackage{algorithmic}
\usepackage{graphicx}
\usepackage{textcomp}
\usepackage{xcolor}
\usepackage{epstopdf}
\usepackage{multirow}
\usepackage{color}

\def\BibTeX{{\rm B\kern-.05em{\sc i\kern-.025em b}\kern-.08em
    T\kern-.1667em\lower.7ex\hbox{E}\kern-.125emX}}
\begin{document}

\title{Efficient Fine-Tuning with Domain Adaptation for Privacy-Preserving Vision Transformer
}

\author{\IEEEauthorblockN{1\textsuperscript{st} Teru Nagamori}
\IEEEauthorblockA{Tokyo Metropolitan University\\
Tokyo,Japan \\
nagamori-teru@ed.tmu.ac.jp}
\and
\IEEEauthorblockN{2\textsuperscript{nd} Sayaka Shiota}
\IEEEauthorblockA{Tokyo Metropolitan University\\
Tokyo,Japan  \\
sayaka@tmu.ac.jp}
\and 
\IEEEauthorblockN{3\textsuperscript{rd} Hitoshi Kiya}
\IEEEauthorblockA{Tokyo Metropolitan University\\
Tokyo,Japan \\
kiya@tmu.ac.jp}

}

\maketitle

\begin{abstract}
We propose a novel method for privacy-preserving deep neural networks (DNNs) with the Vision Transformer (ViT). The method allows us not only to train models and test with visually protected images but to also avoid the performance degradation caused from the use of encrypted images, whereas conventional methods cannot avoid the influence of image encryption. A domain adaptation method is used to efficiently fine-tune ViT with encrypted images. In experiments, the method is demonstrated to outperform conventional methods in an image classification task on the CIFAR-10 and ImageNet datasets in terms of classification accuracy.  
\end{abstract}

\begin{IEEEkeywords}
privacy-preserving, domain adaptation, image classification, Vision Transformer
\end{IEEEkeywords}

\section{Introduction}
Deep neural networks (DNNs) have been deployed in many applications including security-critical ones such as biometric authentication and medical image analysis. In addition, training a deep learning model requires a huge amount of data and fast computing resources, so cloud environments are increasingly used in various applications of DNN models. However, since cloud providers are not always reliable in general, privacy-preserving deep learning has become an urgent problem \cite{kiya2022overview, Encryption-Then-Compression, PFL}.

One of the privacy-preserving solutions for DNNs is to use encrypted images to protect visual information in images for training and testing models \cite{Ito_access, Maung_ICIP}. In this approach, images are transformed by using a learnable encryption method, and encrypted images are used as training and testing data. The use of models trained with encrypted images enables us to use state-of-the-art learning algorithms without any modification. However, all conventional learnable encryption methods \cite{LE, Pixel-Based, madono2020, maung_privacy, qi-san, qi2023colorneuracrypt} have a problem, that is, the performance of encrypted models degrades compared with models without encryption.

Traditional cryptographic methods such as homomorphic encryption \cite{Homomorphic1, Homomorphic2} are one of the other privacy-preserving approaches, but the computation and memory costs are expensive, and it is not easy to apply these methods to state-of-the-art DNNs directly. Federated learning \cite{PFL, FL} allows users to train a global model without centralizing the training data on one machine, but it cannot protect privacy during inference for test data when a model is deployed in an untrusted cloud server.

Accordingly, we focus on the problem of degraded model performance that occurs when models are trained with encrypted images. In this paper, we consider reducing the performance degradation of encrypted models when using encrypted images in the Vision Transformer (ViT) \cite{ViT}, which has high performance in image classification, and we propose a domain adaptation method to reduce the influence of encryption. In experiments, the proposed method is demonstrated to maintain almost the same performance as models trained with plain images in terms of image classification accuracy even when using encrypted images.

The rest of this paper is structured as follows. Section \ref{Related work} presents related work on image encryption for deep learning and the Vision Transformer. Regarding the proposed method, Section \ref{Proposed Method} includes an overview, image encryption, and fine-tuning with domain adaptation. Experiments for verifying the effectiveness of the method, including classification performance, training efficiency, and comparison with conventional methods, are presented in Section \ref{Experiments}, and Section \ref{Conclusion} concludes this paper.

\begin{figure*}[tb]
    \centering
    \includegraphics[bb=0 0 954 644,scale=0.37]{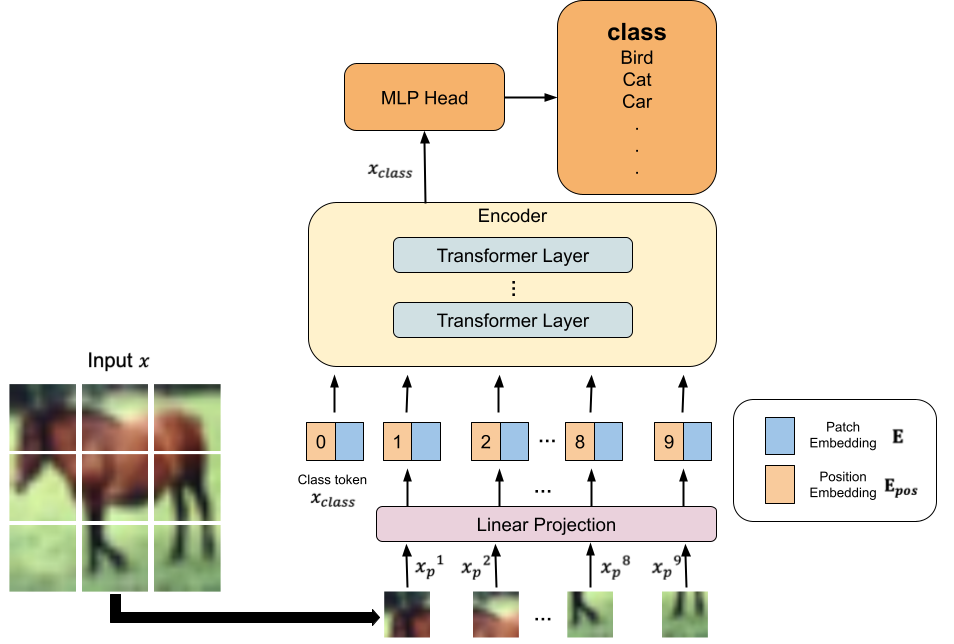}
    \caption{Architecture of Vision Transformer ($N$=9)}
    \label{ViT}
\end{figure*}

\section{Related work}\label{Related work}
\subsection{Image Encryption for Deep Learning}
Various image transformation methods with a secret key, often referred to as perceptual image encryption or image cryptography, have been studied so far for many applications \cite{kiya2022overview}. In this paper, we focus on learnable images transformed with a secret key, which have been studied for deep learning. Learnable encryption enables us to directly apply encrypted data to a model as training and testing data. Encrypted images have no visual information on plain images in general, so privacy-preserving learning can be carried out by using visually protected images. In addition, the use of a secret key allows us to embed unique features controlled with the key into images. Adversarial defenses \cite{maung_AD} and access control \cite{KIYA20232022} are carried out with encrypted data using unique features.

Tanaka first introduced a block-wise learnable image encryption method (LE) with an adaptation layer \cite{LE}, which is used prior to a classifier to reduce the influence of image encryption. Another encryption method is a pixel-wise encryption (PE) method in which negative-positive transformation (NP) and color component shuffling are applied without using any adaptation layer \cite{Pixel-Based}. However, both encryption methods are not robust enough against ciphertext-only attacks, as reported in \cite{chang2020attacks, Ito_access}. To enhance the security of encryption, LE was extended to an extended learnable image encryption method (ELE) by adding a block scrambling (permutation) step and a pixel encryption operation with multiple keys \cite{madono2020}. However, ELE still has an inferior accuracy compared with using plain images, even when an additional adaptation network (denoted as ELE-AdaptNet hereinafter) is applied to reduce the influence of the encryption \cite{madono2020}. Moreover, large-size images cannot be applied to ELE because of the high computation cost of ELE-AdaptNet.

Recently, block-wise encryption was also pointed out to have a high similarity with isotropic networks such as ViT and ConvMixer \cite{jimaging8090233}, and the similarity enables us to reduce performance degradation \cite{maung_privacy, qi-san, jimaging8090233}. In addition, neural network-based image encryption methods without any block-wise encryption, called NeuraCrypt \cite{yala2021neuracrypt} and Color-NeuraCrypt \cite{qi2023colorneuracrypt}, have been proposed, but these methods also have the same performance degradation problem as conventional block-wise encryption methods.

Accordingly, we propose a domain adaptation method so that the performance degradation problem is solved by combining block-wise encryption with domain adaptation under the use of ViT.

\subsection{Vision Transformer}
The Vision Transformer (ViT) \cite{ViT} is commonly used in image classification tasks and is known to provide a high classification performance. As shown in Fig. \ref{ViT}, in ViT, an input image $x \in \mathbb{R}^{h \times w \times c}$ is divided into $N$ patches with a size of $p \times p$, where $h$, $w$, and $c$ are the height, width, and number of channels of the image. Also, $N$ is given as $hw/p^2$. Afterward, each patch is flattened into $x_{p}^i = [x_{p}^i(1), x_{p}^i(2),\dots , x_{p}^i(L)]$. Finally, a sequence of embedded patches is given as
\begin{align} \label{original_input}
    z_{0} =& [x_{class}; x_{p}^1\mathbf{E}; x_{p}^2\mathbf{E}; \dots x_{p}^i\mathbf{E}; \dots x_{p}^N\mathbf{E}] + \mathbf{E_{pos}},
\end{align}
where 
\begin{align*}
\mathbf{E_{pos}} =& ((e_{pos}^0) (e_{pos}^1) \dots (e_{pos}^i) \dots (e_{pos}^N)),\\
  L =& p^{2}c, x_{class} \in \mathbb{R}^{D}, \ x_{p}^{i} \in \mathbb{R}^{L}, \ e_{pos}^i \in \mathbb{R}^{D},\\
  \mathbf{E} \in& \mathbb{R}^{L \times D}, \ \mathbf{E_{pos}} \in \mathbb{R}^{(N+1) \times D}.
\end{align*}
$x_{class}$ is a class token, $\mathbf{E}$ is an embedding (patch embedding) that linearly maps each patch to dimensions $D$, $\mathbf{E_{pos}}$ is an embedding (position embedding) that gives position information to patches in the image, $e_{pos}^0$ is the position information of a class token, $e_{pos}^i$ is the position information of each patch, and $z_0$ is a sequence of embedded patches. Afterward, $z_0$ is input into the transformer encoder. The encoder outputs only the class token, which is a vector of condensed information on the entire image, and it is used for classification. 

Previous studies have indicated that when DNN models are trained with encrypted images, the performance of the models is degraded compared with models trained with plain images. In contrast, it has been pointed out that block-wise encryption has similarity with models that have embedding structures such as ViT \cite{maung_privacy,jimaging8090233}. Accordingly, we focus on the relationship between block-wise image encryption and the embedding structure of ViT to avoid the influence of image encryption even when encrypted images are used for training models.

\begin{figure*}[tb]
    \hspace{70pt}
    \includegraphics[bb=0 0 1500 1045,scale=0.35]{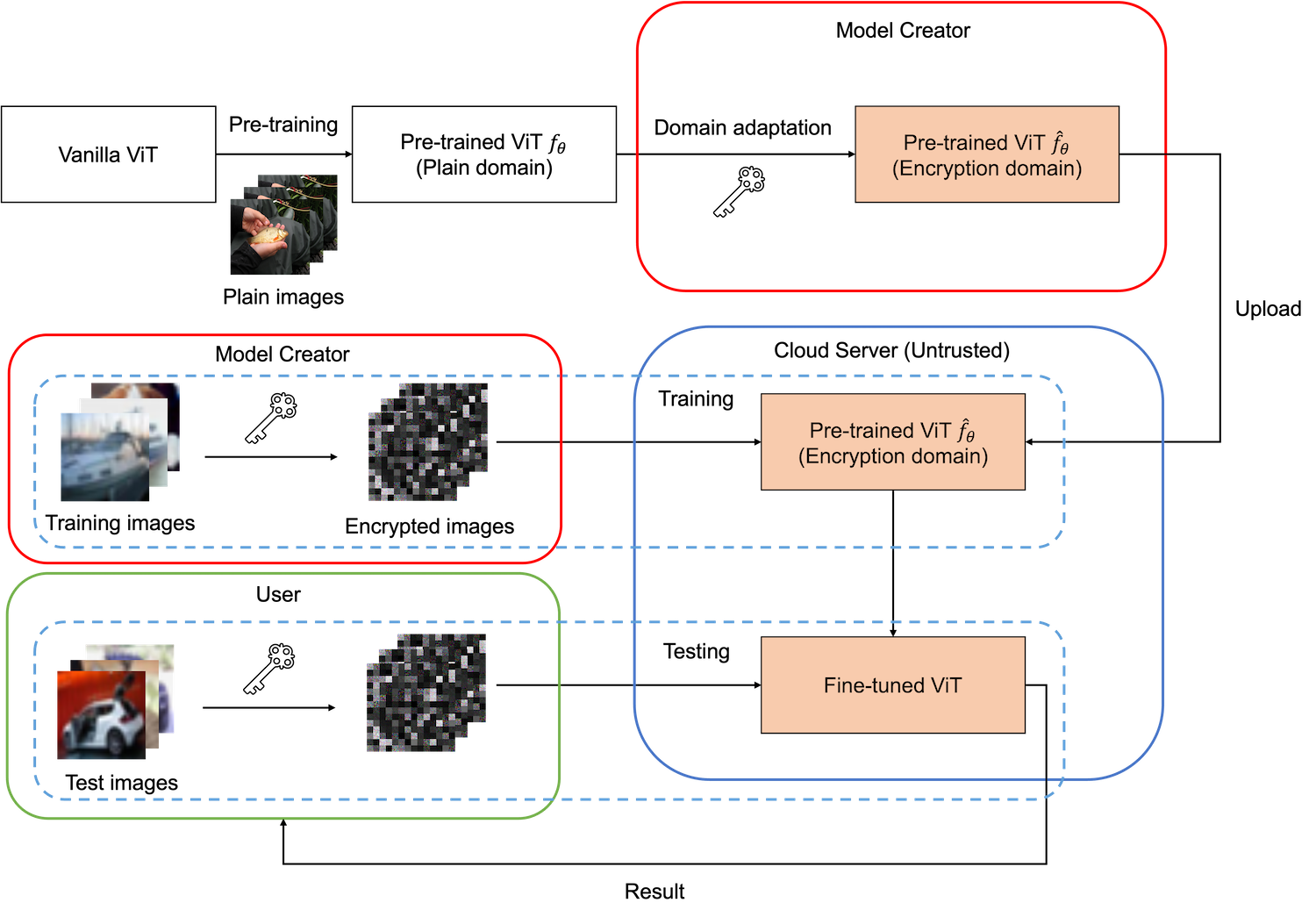}
    \caption{Overview of proposed fine-tuning}
    \label{Overview of DA}
\end{figure*}

\section{Proposed Method}\label{Proposed Method}
\subsection{Overview}
Figure \ref{Overview of DA} shows the procedure for fine-tuning a model by using the proposed domain adaptation. Since a model pre-trained with plain images as in \cite{madono2020} is prepared in advance, the fine-tuning with encrypted images that we will discuss is highlighted in the figure.

A model creator applies the domain adaptation to the pre-trained model $f_{\theta}$ to generate a model $\hat{f}_{\theta}$ with a key. Model $\hat{f}_{\theta}$ is then uploaded to a cloud service such as Amazon Web Services (AWS), which may be untrusted. Model $\hat{f}_{\theta}$ is fine-tuned by using images encrypted with the key. In this paper, we use a block-wise encryption method, which consists of block scrambling and pixel shuffling, to generate encrypted images as an example of block-wise encryption \cite{kiya2022overview}. In block scrambling, an image is divided into non-overlapped blocks (patches), and the divided blocks are randomly permutated with a key. In contrast, pixel shuffling is used to randomly permutate the position of pixels in each block as described later. 

Finally, test images are encrypted with the same keys as those used for fine-tuning by a user, and the encrypted test images are input to the fine-tuned model.

\begin{figure*}[tb]
    \centering
    \includegraphics[bb=0 0 1084 522,scale=0.4]{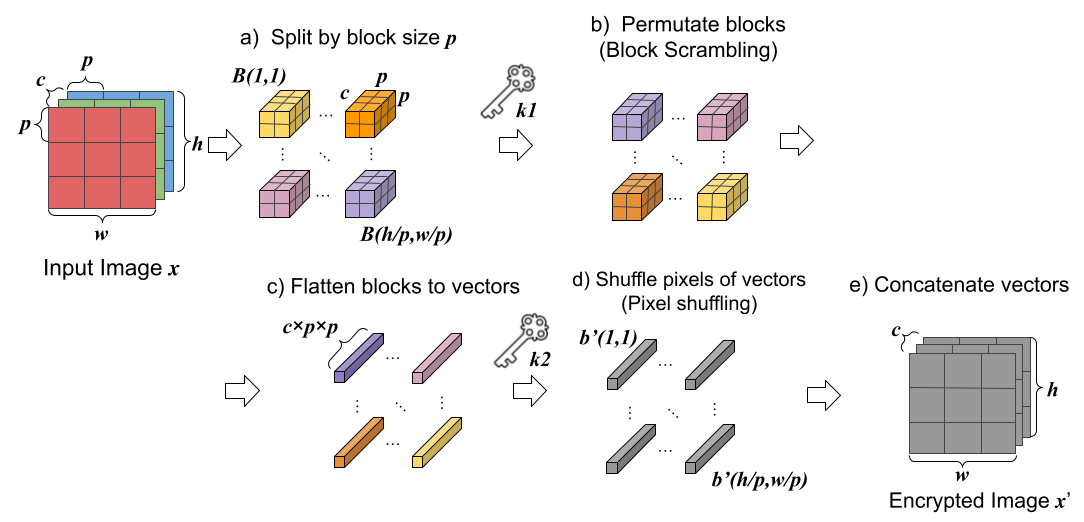}
    \caption{Procedure of image encryption}
    \label{Image Encryption}
\end{figure*}

\subsection{Image Encryption}
Model $f_{\theta}$ is fine-tuned by using encrypted images as shown in Fig. \ref{Overview of DA}. Figure \ref{Image Encryption} shows the encryption procedure used in this paper. Details on the procedure are given below.

\begin{itemize}
\item[(a)] Divide an image $x \in \mathbb{R}^{h \times w \times c}$ into non-overlapped blocks with a size of $p \times p$ such that $B = (\left\{B_{1},\dots, B_{i} ,\dots, B_{N}\right\})^\top $, $B_{i} \in \mathbb{R}^{p^2 \times c} $.

\item[(b)] Permutate blocks with secret key $k_1$ such as
        \begin{align} \label{eq: bs}
                    \hat{B} = &  \mathbf{E_{bs}}B \\
                    = & (\left\{\hat{B}_{1},\dots, \hat{B}_{i} ,\dots ,\hat{B}_{N}\right\})^\top,  \hat{B}_{i} \in \mathbb{R}^{p^2 \times c}  \notag
        \end{align}
where $\hat{B}_{i} \in \left\{B_{1},\dots ,B_{N}\right\}$, and a permutation matrix $\mathbf{E_{bs}}$ is generated with key $k_1$ as follows.

    \begin{itemize}
            \item[1)]Using key $k_1$, generate a random integer vector with a length of $N$ as 
                            \begin{equation}
                                l_{e} = [l_{e}(1), l_{e}(2), \dots, l_{e}(i)\ , \dots ,l_{e}(N)] \ ,
                            \end{equation}
            where
                            \begin{align*}
                                l_{e}(i) \in& \left\{1,2,...,N\right\}, \\
                                l_{e}(i) \neq& l_{e}(j) \ \text{if} \ i \neq j, \\
                                i,j \in& \left\{1, \dots, N\right\}.
                            \end{align*}
            \item[2)] $k_{(i,j)}$ is given as
                \begin{equation}
                    k_{(i,j)} = \left\{ \begin{matrix}0&(j\neq l_{e}(i))\\ 1&(j = l_{e}(i)) \end{matrix} \right. .
                \end{equation}

            \item[3)] Define $\mathbf{E_{bs}} \in \mathbb{R}^{N \times N}$ as
                \begin{align} \label{eq: ebs}
                    \mathbf{E_{bs}} =
                    \begin{bmatrix} 
                          k_{(1,1)} & k_{(1,2)} & \dots & k_{(1,N)} \\
                          k_{(2,1)} & k_{(2,2)} &\dots  & k_{(2,N)} \\
                          \vdots & \vdots & \ddots & \vdots \\
                          k_{(N,1)} & k_{(N,2)} & \dots & k_{(N,N)}\\
                    \end{bmatrix},
                \end{align}
        \end{itemize}
For example, $\mathbf{E_{bs}}$ is given by, for $N = 3$,
        \begin{align}
            \mathbf{E_{bs}} =
                \begin{bmatrix} 
                      1 & 0 & 0 \\
                      0 & 0 & 1 \\          
                      0 & 1 & 0 \\
                \end{bmatrix}.
        \end{align}
\item[(c)]Flatten each block $\hat{B_{i}}$ into a vector $\hat{b}_{i}$ with a length of $L = p \times p \times c$ such that
        \begin{equation}
            \hat{b}_{i} = [\hat{b}_{i}(1), \dots , \hat{b}_{i}(L)]^\top,
        \end{equation}
where $\hat{b}_{i}$ is the same as $(x_{p}^i)^\top$ in Eq. \ref{original_input}.

\item[(d)]Shuffle pixels in $\hat{b}_{i}$ with secret key $k_2$ such as
        \begin{align} \label{eq: 8}
                    b'_{i} =& (\mathbf{E}_{ps}\hat{b}_{i})^\top \\ \notag
                    =& [b'_{i}(1), \dots , b'_{i}(L)],
        \end{align}
where $\mathbf{E_{ps}}$ is a matrix generated by the same procedure as $\mathbf{E_{bs}}$ ($N$ is replaced with $L$).

Also, Eq. \ref{eq: 8} is expressed as
        \begin{equation} \label{eq: ps}
                    b'_{i} = (\mathbf{E}_{ps}\hat{b}_{i})^\top = (\mathbf{E}_{ps}(x_{p}^{i})^\top)^\top = x^{'i}_{p}.
        \end{equation}

\item[(e)]Concatenate the encrypted vectors $b'_{i} (i \in \left\{1, \dots, N\right\})$ into an encrypted image $x^{'}$.
\end{itemize}

\begin{figure*}[tb]
    \centering
    \scalebox{0.5}[0.5]{
    \begin{tabular}{cccc}
        \begin{minipage}[b]{0.4\hsize}
          \centering
          \includegraphics[bb=0 0 224 224,scale=0.6]{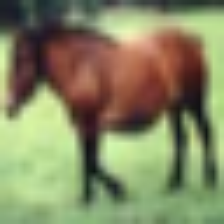}
        \end{minipage}
        &
        \begin{minipage}[b]{0.4\hsize}
          \centering
          \includegraphics[bb=0 0 224 224,scale=0.6]{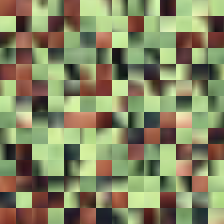}
        \end{minipage}
        &
        \begin{minipage}[b]{0.4\hsize}
          \centering
          \includegraphics[bb=0 0 224 224,scale=0.6]{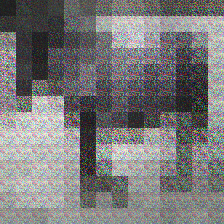}
        \end{minipage}
        &
        \begin{minipage}[b]{0.4\hsize}
          \centering
          \includegraphics[bb=0 0 224 224,scale=0.6]{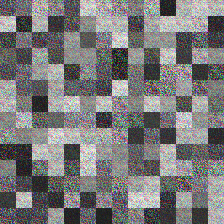}
        \end{minipage} \\
        
        \begin{tabular}{c}
    (a) Original \\ (224 $\times$ 224 $\times$ 3)
            \end{tabular}
            &
            \begin{tabular}{c}
    (b) Block scrambling \\ (block size = 16)
            \end{tabular}
            &
            \begin{tabular}{c}
    (c) Pixel shuffling \\ (block size = 16)
            \end{tabular}
            &
            \begin{tabular}{c}
    (d) Block scrambling + Pixel shuffling \\ (block size = 16)
            \end{tabular}
        \end{tabular}
        }
    \caption{Example of encrypted images}
    \label{Encrypted Images}
\end{figure*}

Figure \ref{Encrypted Images} shows an example of images encrypted with this procedure, where block size $p$ was 16.

\subsection{Fine-tuning with Domain Adaptation}
As shown in Fig. \ref{Overview of DA}, model $f_{\theta}$ pre-trained in the plain domain is transformed into model $\hat{f}_{\theta}$ to reduce the influence of encryption prior to the fine-tuning of $f_{\theta}$ by using the proposed domain adaptation. The procedure of the adaptation is summarized here.

The domain adaptation is carried out in accordance with the embedding structure of ViT. $\mathbf{E_{bs}}$ in Eq. \ref{eq: ebs} is used to permutate blocks in an image, so it has a close relationship with position embedding $\mathbf{E_{pos}}$ in Eq. \ref{original_input}, where $\mathbf{E_{pos}}$ includes the position information of a class token $x_{class}$. In contrast, $\mathbf{E_{bs}}$ does not consider the information of $x_{class}$. To fill the gap between $\mathbf{E_{pos}}$ and $\mathbf{E_{bs}}$, $\mathbf{E_{bs}}$ is extended as 

\begin{equation}
  \mathbf{E_{bs}'} =
    \begin{bmatrix} 
          1 & 0 & 0 & \dots & 0 \\
          0 & k_{(1,1)} & k_{(1,2)} & \dots & k_{(1,N)} \\
          0 & k_{(2,1)} & k_{(2,2)} &\dots  & k_{(2,N)} \\
          \vdots &\vdots & \vdots & \ddots & \vdots \\
          0 & k_{(N,1)} & k_{(N,2)} & \dots & k_{(N,N)}\\
    \end{bmatrix},
\end{equation}
\begin{align*}
           \mathbf{E_{bs}'} &\in \mathbb{R}^{(N + 1) \times (N + 1)}.
\end{align*}

In the domain adaptation, $\mathbf{E_{pos}}$ is transformed as in

\begin{equation}\label{eq: pos}
    \mathbf{\hat{E}_{pos}} =\mathbf{E_{bs}'}\mathbf{E_{pos}}.
\end{equation}

By using Eq. \ref{eq: pos}, $\mathbf{E_{pos}}$ can be adapted to images permutated by block scrambling. Similarly, $\mathbf{E}$ is adapted to pixels that are randomly replaced by pixel shuffling, as in

\begin{equation}\label{eq: patch}
    \mathbf{\hat{E}} =\mathbf{E_{ps}}\mathbf{E}.
\end{equation}

From Eqs. \ref{eq: bs}, \ref{eq: ps}, \ref{eq: pos}, and \ref{eq: patch}, under the use of image encryption and domain adaptation, an adapted sequence of embedded patches in Eq. \ref{original_input} is replaced with

\begin{align} \label{eq: enc input}
    \hat{z}_{0} =& \mathbf{E_{bs}'}[x_{class}; (\mathbf{E_{ps}}(x_{p}^1)^\top)^\top\mathbf{E_{ps}}\mathbf{E} ; \dots (\mathbf{E_{ps}}(x_{p}^N)^\top)^\top\mathbf{E_{ps}}\mathbf{E}]\\ \notag
    +& \mathbf{E_{bs}'}\mathbf{E_{pos}}\\ \notag
    =& \mathbf{E_{bs}'}[x_{class}; x_{p}^{'1}\mathbf{\hat{E}}; \dots x_{p}^{'N}\mathbf{\hat{E}}] + \mathbf{E_{bs}'}\mathbf{E_{pos}}.
\end{align}\\

In accordance with Eqs. \ref{eq: pos} and \ref{eq: patch}, $f_\theta$ with $\mathbf{E_{pos}}$ and $\mathbf{E}$ is transformed to $\hat{f}_\theta$ with $\mathbf{\hat{E}_{pos}}$ and $\mathbf{\hat{E}}$ as shown in Fig. \ref{Overview of DA}, where the difference between $f_\theta$ and $\hat{f}_\theta$ is only a sequence of embedding patches $\hat{z}_{0}$ in Eq. \ref{eq: enc input}.

\begin{figure*}[tb]
    \hspace{40pt}
    \includegraphics[bb=0 0 1000 228,scale=0.6]{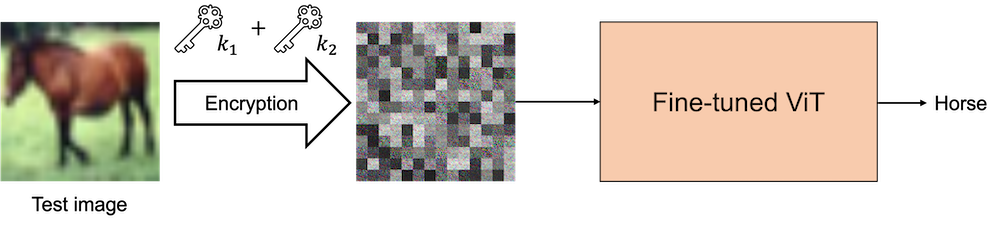}
    \caption{Test procedure}
    \label{testflow}
\end{figure*}

Finally, test images are applied to $\hat{f}_\theta$ in the adaptation domain. In our framework, as shown Fig. \ref{testflow}, test images are input to a model after fine-tuning $\hat{f}_\theta$ with encrypted images. In addition, test images are required to be encrypted by using the same keys as those used for fine-tuning the model.

\section{Experiments}\label{Experiments}
The effectiveness of the proposed method is verified in terms of the accuracy of image classification and the efficiency of model training.

\subsection{Setup}
We conducted image classification experiments on the CIFAR-10 \cite{cifar10} and CIFAR-100 datasets \cite{cifar10}. CIFAR-10 consists of 60,000 images with 10 classes (6000 images for each class), and CIFAR-100 consists of 60,000 images with 100 classes (600 images for each class), where 50,000 images were used for training, and 10,000 images were used for testing. In addition, we used the Imagenette dataset \cite{imagenette}, which is a subset of the ImageNet dataset with 10 classes. We also fine-tuned a ViT-B\underline{ }16 model pre-trained with the ImageNet21k dataset, so we resized the image size from $32 \times 32 \times 3$ to $224 \times 224 \times 3$. Training was carried out under the use of a batch size of 32, a learning rate of 0.001, a momentum of 0.9, and a weight decay of 0.0005 using the stochastic gradient descent (SGD) algorithm for 15 epochs. A cross-entropy loss function was used as the loss function. The block size of the encryption was set to 16 to match the ViT patch size.

\begin{table*}[tb]
\caption{Classification accuracy (\%) with and without proposed method (CIFAR-10)}
 \label{table:cifar10}
 \centering
  \begin{tabular}{c|cc}
   \hline
    & \multicolumn{2}{c}{Domain Adaptation} \\
    Encryption & Yes & No \\
    \hline
   Pixel shuffling & \textbf{99.05} & 97.78 \\
   Block scrambling & \textbf{98.93} & 94.25\\
   Pixel shuffling + Block scrambling & \textbf{98.98} & 72.86\\
   \hline
   Plain (Baseline)& \multicolumn{2}{c}{99.00}\\
   \hline
  \end{tabular}
\end{table*}

\begin{figure*}[tb]
\centering
    \begin{tabular}{ccc}
        \hspace{-30pt}
        \begin{minipage}{4truecm}
            \centering
              \includegraphics[bb=0 0 1140 860,scale=0.16]{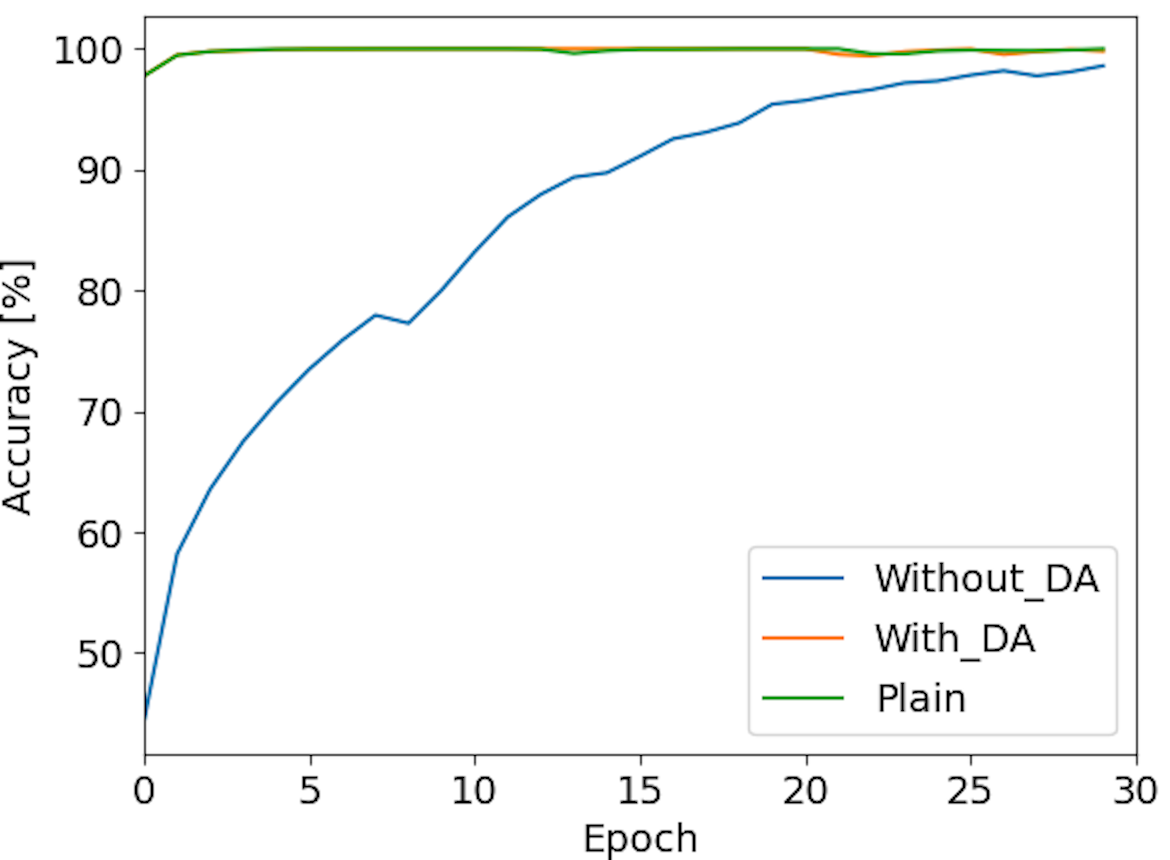}
        \end{minipage}
        & 
        \hspace{20pt}
        &
        \begin{minipage}{4truecm}
            \centering
              \includegraphics[bb=0 0 1140 860,scale=0.16]{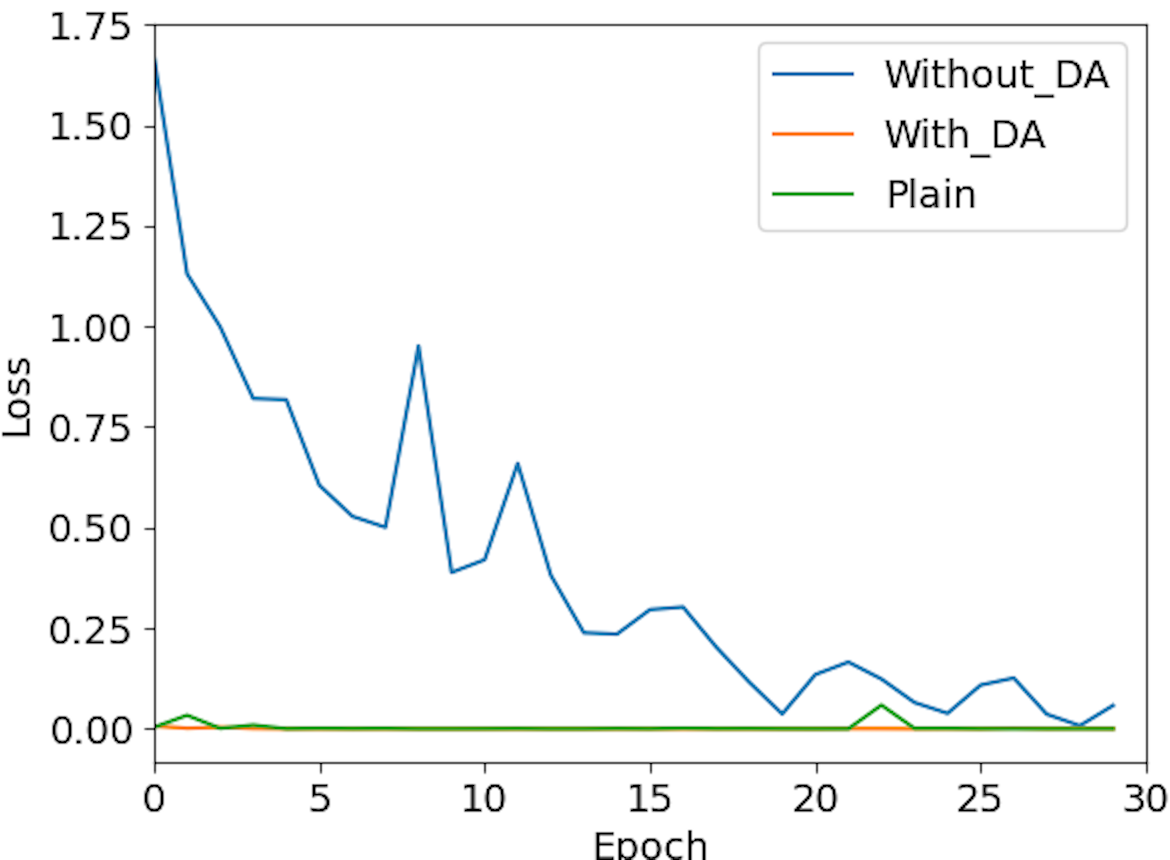}
        \end{minipage}
        \\
\multicolumn{3}{c}{\scriptsize (a)Training} \vspace{-30pt}
        \\
        \hspace{-30pt}
        \begin{minipage}{4truecm}
            \centering
            \includegraphics[bb=0 0 1140 860,scale=0.16]{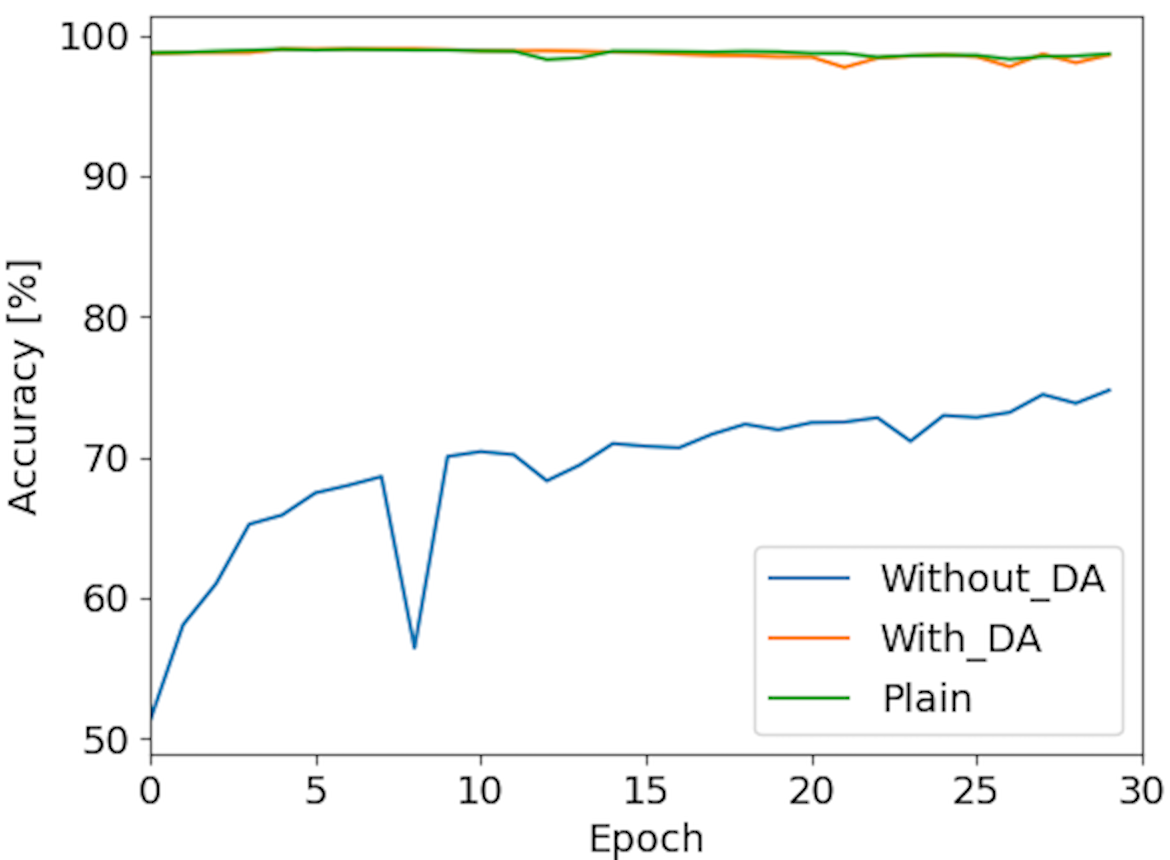}
            \end{minipage}
        &
        \hspace{20pt}
        &
        \begin{minipage}{4truecm}
            \centering
             \includegraphics[bb=0 0 1140 860,scale=0.16]{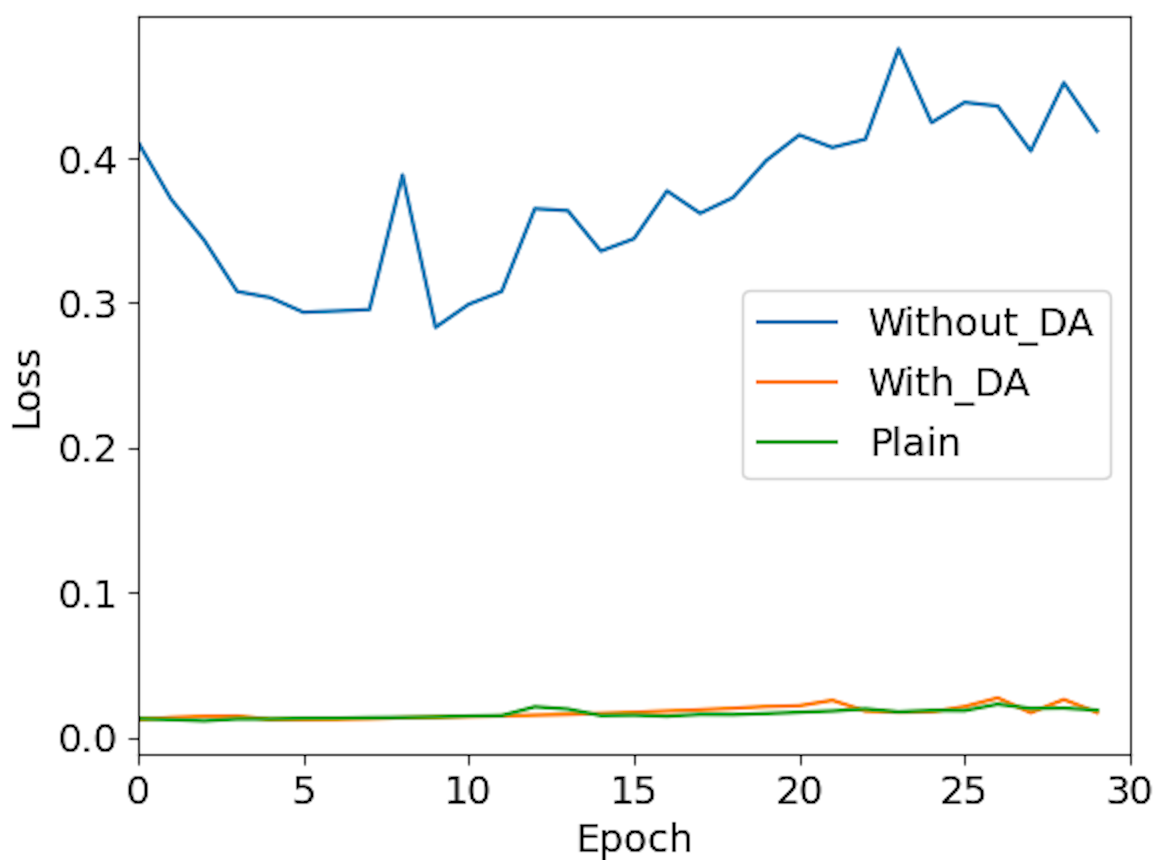}
        \end{minipage}
        \\
\multicolumn{3}{c}{\scriptsize (b)Testing}
        \\
    \end{tabular}
\caption{Learning curves during training and testing}
    \label{Training efficiency}
\end{figure*}

\subsection{Classification Performance}
Table \ref{table:cifar10} shows the classification accuracy of models trained with encrypted images. When training models without the proposed domain adaptation, the accuracy decreased compared with that of using plain images. In particular, when applying the combined use of pixel shuffling and block scrambling, the classification accuracy significantly decreased due to the influence of encryption. In contrast, when training models with the domain adaptation, the accuracy was almost the same as that of using plain images (Baseline). Accordingly, we confirmed that the proposed domain adaptation was effective in fine-tuning pre-trained models with encrypted images.

\subsection{Training Efficiency}
When training a model, it is important to consider not only the classification performance but also the training efficiency. Training efficiency means the time required for model training and loss convergence. The use of encrypted images should be avoided as it significantly increases the time required for training and loss convergence compared with models trained using plain images.

Figure \ref{Training efficiency} shows the classification accuracy and loss value of three models per epoch during training and testing where the number of epochs was increased to 30. For training, from the figure, a model fine-tuned with only encrypted images (without\underline{ }DA) needed a larger number of epochs to achieve the same values as those of the plain model (Plain). In contrast, when using the domain adaptation, the model fine-tuned with encrypted images (Proposed) achieved almost the same performances in each epoch as the model trained with plain images. This means that even if encrypted images are used, the domain adaptation allows us not only to train privacy-preserving models with almost the same accuracy as plain models but to also avoid increasing the model training time.

For testing, the model without the domain adaptation was confirmed to have degraded accuracy and loss performances compared with the plain model. In contrast, the model with domain adaptation had almost the same performances as the plain model.

Accordingly, the proposed method allows us not only to train privacy-preserving models without performance degradation but to also avoid an increase in training time even when using encrypted images.

\begin{table*}[tb]
 \caption{Comparison with existing methods (CIFAR-10 and CIFAR-100)}
 \label{table:comparison}
 \centering
  \begin{tabular}{cccc}
   \hline
    \multirow{2}{*}{Encryption} & \multirow{2}{*}{Model} & \multicolumn{2}{c}{Accuracy (\%) $\uparrow$} \\
        &    &  CIFAR-10 & CIFAR-100 \\
    \hline
    Block scrambling $+$ Pixel shuffling $+$ Domain adaptation (Proposed) & ViT-B\underline{ }16 & \textbf{98.98} & \textbf{92.58}\\
   LE \cite{LE} & Shakedrop $+$ AdaptNet  & 94.49 & 75.48 \\
   ELE \cite{madono2020} &  Shakedrop $+$ AdaptNet & 83.06 & 62.97 \\
   PE \cite{Pixel-Based} & ResNet-18 & 92.03 & - \\
   Block scrambling $+$ Pixel shuffling $+$ Negative/Positive transformation \cite{jimaging9040085} &  ConvMixer-512/16 $+$ Adaptive matrix & 92.65 & -\\
   Color-NeuraCrypt \cite{qi2023colorneuracrypt} & ViT-B\underline{ }16 & 96.20 & -\\
   Block scrambling $+$ Pixel position shuffling \cite{qi-san} & ViT-B\underline{ }16 & 96.64 & 84.42\\
   EtC \cite{maung_privacy} & ViT-B\underline{ }16 & 87.89 & -\\
    &  ConvMixer-256/8 & 92.72 & -\\
   \hline
   \multirow{4}{*}{Plain} & ShakeDrop & 96.70 & 83.59\\
    & ResNet-18 & 95.53 & -\\
    & ViT-B\underline{ }16 & 99.00 & 92.60\\
    & ConvMixer-256/8 & 96.07 & -\\
   \hline
  \end{tabular}
\end{table*}

\begin{table*}[tb]
 \caption{Comparison with existing methods (Imagenette)}
 \centering
 \label{table:imagenette}
  \begin{tabular}{ccc}
   \hline
    Encryption & Model & Accuracy(\%) $\uparrow$ \\
    \hline
    Block scrambling $+$ Pixel shuffling $+$ Domain adaptation (Proposed) & ViT-B\underline{ }16 & \textbf{99.2} \\
    EtC \cite{maung_privacy} & ViT-B\underline{ }16 & 90.62 \\
    & ConvMixer-384/8 & 90.11 \\
   \hline
   \multirow{2}{*}{Plain} & ViT-B\underline{ }16 & 99.2\\
    & ConvMixer-384/8 & 96.10 \\
   \hline
  \end{tabular}
\end{table*}

\subsection{Comparison with Conventional Methods}
In Tables \ref{table:comparison} and \ref{table:imagenette}, the proposed method is compared with the state-of-the-art methods for privacy preserving image classification with encrypted images such as LE \cite{LE}, PE \cite{Pixel-Based}, and ELE \cite{madono2020}. From the tables, the proposed method not only has the lowest performance degradation on all datasets but also outperforms all existing methods in terms of the image classification accuracy.

\section{Conclusion} 
\label{Conclusion}
In this paper, we proposed a domain adaptation method to reduce the influence of image encryption for privacy-preserving ViT. The method is applied to a model pre-trained in the plain domain prior to the fine-tuning of the pre-trained model. The domain adaptation is carried out with two random matrices generated with secret keys, and test images are also encrypted by using the keys. In experiments, the method was demonstrated not only to reduce the performance degradation of models but to also achieve the highest classification accuracy among conventional methods on the CIFAR-10, CIFAR-100, and Imagenette datasets.

\section*{Acknowledgment}
This study was partially supported by JSPS KAKENHI (Grant Number JP21H01327) and JST CREST (Grant Number JPMJCR20D3).

\bibliographystyle{IEEEtran} 
\bibliography{references} 

\end{document}